\pdfoutput=1
\documentclass{article}

\usepackage[preprint]{neurips_2025}

\usepackage[utf8]{inputenc}
\usepackage[T1]{fontenc}
\usepackage{hyperref}
\usepackage{url}
\usepackage{booktabs}
\usepackage{amsfonts}
\usepackage{amsmath}
\usepackage{microtype}
\usepackage{xcolor}
\usepackage{graphicx}
\usepackage{listings}
\hypersetup{colorlinks=true, linkcolor=blue!50!black, citecolor=blue!50!black, urlcolor=blue!50!black}

\newcommand{\boonta}{\textsc{boonta}}

\lstdefinestyle{py}{
  language=Python,
  basicstyle=\ttfamily\footnotesize,
  breaklines=true,
  columns=fullflexible,
  keywordstyle=\color{blue!60!black},
  commentstyle=\color{gray},
  stringstyle=\color{green!40!black},
  showstringspaces=false,
  frame=single,
  framerule=0.3pt,
  rulecolor=\color{gray!50},
  xleftmargin=2pt,
  xrightmargin=2pt,
  aboveskip=6pt,
  belowskip=6pt,
}
\title{Abstractions for Scaling Reinforcement Learning in JAX}

\author{
  Noah Farr \\
  Technical University of Darmstadt \\
  \texttt{noah.farr@stud.tu-darmstadt.de}
}

\begin{document}

\maketitle

\begin{abstract}
Reinforcement learning (RL) is no longer confined to games and simulated control but has become central to the post-training of large language and vision-language-action models.
We present \boonta{}, a distributed RL library in JAX that unifies these settings under one set of abstractions, which keep the techniques for scaling out of algorithms and environments.
Its training loops, networks and data sources provide multi-GPU training, asynchronous collection, reduced precision and custom kernels, so the same algorithm implementation runs from CartPole to the post-training of a language model.
PPO on Craftax-Classic scales across the GPUs of a node, the recurrent behavior cloning implementation pretrains GPT-2, and recurrent GRPO post-trains Qwen3 to play a number-guessing game through tool calls, all without changing the algorithms. With the same settings, \boonta{} matches the throughput of Stoix on most benchmarks, and its scaling techniques make it faster with large batches.
The code is released under the Apache 2.0 license at \url{https://github.com/noahfarr/boonta}.
\end{abstract}

\section{Introduction}
\label{sec:introduction}

Reinforcement learning (RL) is used at very different scales.
A policy for CartPole trains in seconds on one device, PPO on Craftax trains for a billion environment steps in under an hour on one GPU \citep{matthews2024craftax}, and large language and vision-language-action models are post-trained with policy-gradient methods such as GRPO and PPO \citep{shao2024deepseekmath, li2026simplevla, chen2025pirl}.
As the scale grows, a correct update rule is no longer enough.
Experience has to be split across devices and hosts, slow simulators have to run alongside learning, datasets may exceed device memory, and large or recurrent policies need reduced precision, fast kernels and memory for their recurrent state or key-value cache.

\boonta{} provides these techniques through its components rather than its algorithms.
An algorithm consists of three pure functions, which create its state, choose actions and update the state from a batch of transitions, and it marks the parts of its state that belong to individual environments.
Environments create and advance their own state.
The training loop, called a \emph{podracer}, decides when and where the agent acts and learns.
Anakin and Sebulba implement the synchronous and asynchronous architectures of \citet{hessel2021podracer}, and Quadinaros trains from recorded data.
From the marks on the algorithm state, a podracer partitions rollouts, replay buffers and datasets across the devices of one or several hosts, and the updates keep rollout minibatches and replay samples local to each device.
Networks set their numeric precision, run linear recurrences with Pallas kernels, and treat the key-value cache of a pretrained transformer as recurrent state, so that a language model trains as a recurrent policy with LoRA on quantized weights.

With these components, \boonta{} implements on-policy, off-policy and offline methods, recurrent policies, multi-agent training and curricula such as prioritized level replay and a PSRO league.
It runs environments written in JAX as well as Gymnasium environments and native simulators, and the same implementations train agents from CartPole to the post-training of Qwen3 on a game played through tool calls.

We measure how training scales on one node of eight GPUs.
With a fixed number of environments per GPU, eight GPUs process 4.6 and 7.1 times as many environment steps per second as one GPU for PPO on Breakout-MinAtar and on Craftax-Classic, and 6.5 times for DQN on Breakout-MinAtar.
With a fixed global batch, PPO trains 3.7 times faster on Craftax-Classic and reaches the same returns, and the same PPO implementation reaches similar returns with synchronous and asynchronous training.
Recurrent GRPO post-trains Qwen3-0.6B to play a number-guessing game through tool calls, and the share of games it wins rises from 10\% to 31\% in two hours on one A100.
The appendix gives learning curves for the online and offline methods.

\paragraph{Related work.}
PureJaxRL showed that compiling the environment and the agent into a single program yields large speedups and popularized single-file implementations in JAX \citep{lu2022discovered}.
Rejax \citep{liesen2024rejax} is designed to train many seeds or hyperparameter settings in parallel by vectorizing whole training runs.
\citet{hessel2021podracer} describe two architectures for RL on TPU pods, Anakin for environments in JAX and Sebulba for other environments, the latter in the line of actor-learner designs such as IMPALA \citep{espeholt2018impala} and SEED RL \citep{espeholt2020seed}.
Stoix \citep{toledo2024stoix} and Mava \citep{dekock2023mava} implement both for single-agent and cooperative multi-agent RL, respectively, and Cleanba \citep{huang2023cleanba} made the actor-learner architecture reproducible, a design that the Sebulba podracer of \boonta{} follows.
Component-based frameworks such as RLlib \citep{liang2018rllib}, Acme \citep{hoffman2020acme} and TorchRL \citep{bou2024torchrl} likewise structure RL systems as composable parts.
\boonta{} shares this goal, but attaches methods that adapt the training distribution through a hook that every training loop invokes, and keeps all interfaces, including the hooks, as pure functions, so that Anakin and Quadinaros compile all components into a single program.
EnvPool \citep{weng2022envpool}, Sample Factory \citep{petrenko2020sample} and PufferLib \citep{suarez2024pufferlib} focus on fast simulation and high throughput for environments outside JAX.
TRL \citep{vonwerra2020trl}, verl \citep{sheng2024hybridflow} and OpenRLHF \citep{hu2024openrlhf} are designed for language models and support considerably larger models than \boonta{} currently does.
Table~\ref{tab:comparison} compares how these systems collect experience and distribute computation. Appendix~\ref{app:design} discusses implementation differences.

\begin{figure}[t]
  \centering
  \includegraphics[width=\linewidth]{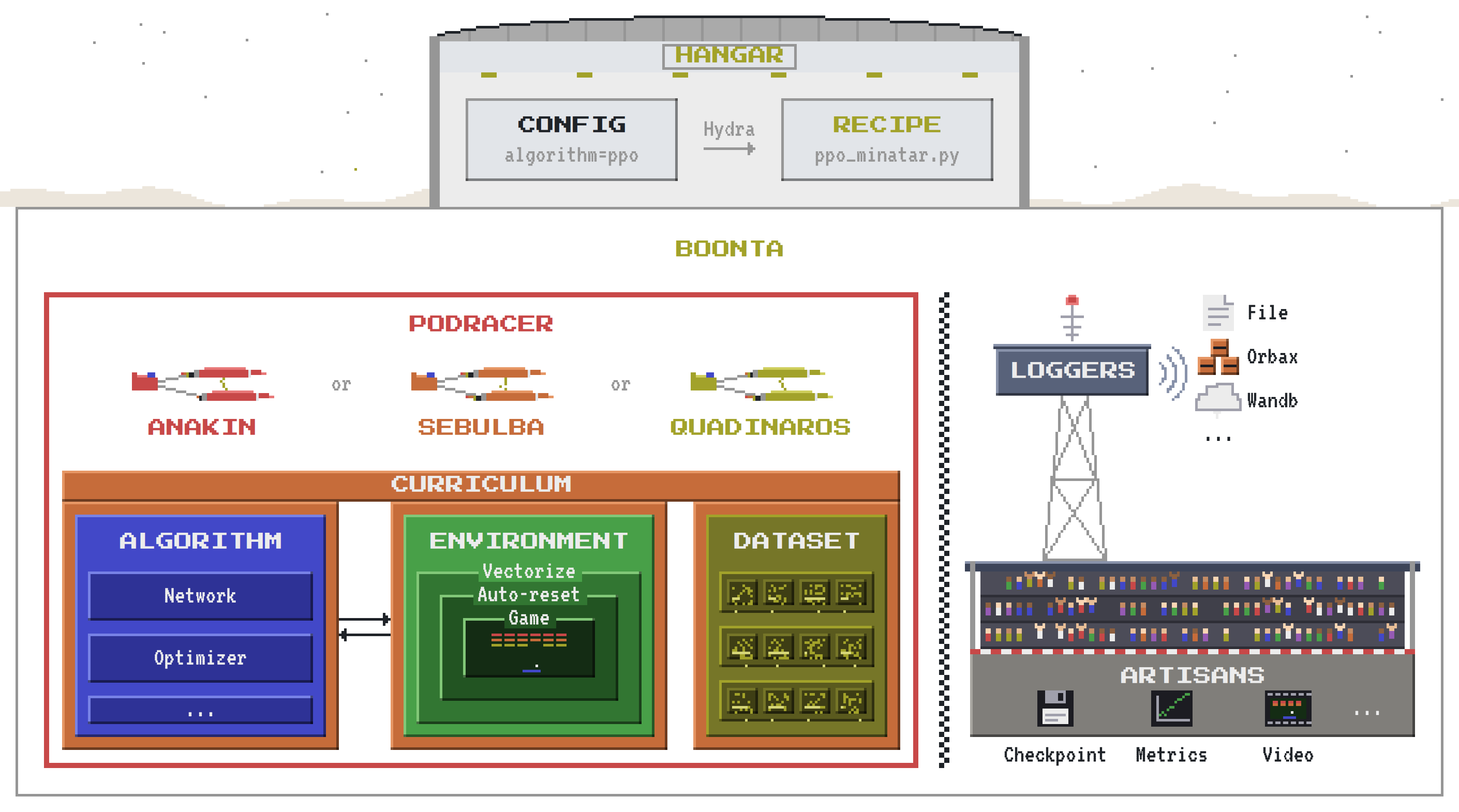}
  \caption{Structure of a \boonta{} run.
  In \texttt{hangar}, a recipe constructs the components from a Hydra configuration and passes them to a podracer. The components below the hangar belong to the library.
  The curriculum wraps the algorithm and the environment and may supply a pit.
  Artisans produce checkpoints, metrics and videos from the podracer state, and loggers write them out.}
  \label{fig:structure}
\end{figure}

\begin{table}[t]
  \centering
  \caption{Scaling strategies in representative training configurations. Parallel seeds train independent policies. Data-parallel learners share one policy update across devices. Library names link to the implementation or documentation for each entry.}
  \label{tab:comparison}
  \small
  \setlength{\tabcolsep}{3pt}
  \renewcommand{\arraystretch}{1.16}
  \begin{tabular}{@{}p{.17\linewidth}p{.35\linewidth}p{.43\linewidth}@{}}
    \toprule
    Library & Experience collection & Parallel execution \\
    \midrule
    \href{https://github.com/noahfarr/boonta}{\boonta{}} &
    Compiled rollouts, asynchronous actors, or batches from disk &
    Anakin and Quadinaros shard data across GPUs and hosts. Sebulba assigns acting and learning to devices on one host. Policy weights are replicated. \\
    \href{https://github.com/luchris429/purejaxrl}{PureJaxRL}, \href{https://github.com/keraJLi/rejax}{Rejax} &
    JAX environments compiled with the learner &
    Vectorize whole training runs over seeds or hyperparameters. Rejax also maps independent runs across devices. \\
    \href{https://github.com/EdanToledo/Stoix}{Stoix}, \href{https://github.com/instadeepai/Mava}{Mava} &
    Compiled JAX rollouts (Anakin) or separate actors (Sebulba) &
    Anakin runs the learner across devices. Sebulba assigns devices to acting and learning for supported algorithms. \\
    \href{https://github.com/vwxyzjn/cleanba}{Cleanba} &
    Actor threads feed a learner with fixed policy lag &
    Actor and learner GPU assignments are configurable. The learner can span several devices. \\
    \href{https://github.com/facebookresearch/minimax}{minimax} &
    Concurrent rollouts on new and replayed levels &
    Distributes curriculum training across devices and can evaluate new levels alongside replay. \\
    \href{https://github.com/PufferAI/PufferLib}{PufferLib} &
    Rollout workers use separate CUDA streams &
    CUDA graphs replay learner updates, and multi-GPU runs synchronize gradients with NCCL. \\
    \href{https://docs.pytorch.org/rl/stable/reference/collectors.html}{TorchRL} &
    Synchronous or asynchronous collectors &
    Collection workers can span machines using Ray or RPC. The training script selects the learner independently. \\
    \href{https://docs.ray.io/en/latest/rllib/scaling-guide.html}{RLlib} &
    Vectorized environments per rollout worker &
    The number of rollout workers and data-parallel learners can be varied independently on a Ray cluster. \\
    \href{https://github.com/google-deepmind/acme/blob/master/acme/agents/tf/d4pg/agent_distributed.py}{Acme (D4PG)} &
    Actors send transitions to Reverb &
    Actors, replay and learner run as separate services. The learner can replicate its update across accelerators. \\
    \href{https://verl.readthedocs.io/en/latest/hybrid_flow.html}{verl} &
    Inference workers generate token sequences &
    Training workers can shard the policy with FSDP or use tensor and pipeline parallelism with Megatron. \\
    \bottomrule
  \end{tabular}
\end{table}

\section{Training architectures}
\label{sec:design}

A podracer receives an algorithm and a batched environment. It initializes their states, supplies batches for the algorithm update and evaluates the policy. A recipe in \texttt{hangar} constructs the same objects from a configuration (Figure~\ref{fig:structure}).

\paragraph{Synchronous training.}
Anakin compiles policy inference, environment stepping and the learner update into one JAX program. Each device advances its own batch of environments for a fixed rollout length, then participates in the update. Nested scans run the environment steps and the updates. The compiled training call donates its input state, allowing XLA to reuse eligible buffers for the next state.

\paragraph{Asynchronous training.}
Sebulba runs rollout collection and learner updates concurrently. Actors can share a device with the learner or run on dedicated devices. With separate devices, actors can simulate environments and run policy inference while the learner updates on the previous batch. The relative cost of these stages determines how much collection can overlap an update.
Each actor sends a rollout through a queue. The learner takes one rollout from each actor per update, so a slow actor can delay the next batch. Following \citet{huang2023cleanba}, the schedule fixes the policy lag at one update. In steady state, actors collect rollout $k$ with parameters from update $k-2$ while the learner computes update $k-1$. Appendix~\ref{app:example} shows how the same PPO implementation runs with either architecture.

\paragraph{Offline training.}
Quadinaros samples recorded transitions in a compiled update loop. BC fits the recorded actions directly. IQL uses the same transitions to fit value and action-value functions and to weight the policy update. Both use the environment for evaluation. A dataset can refresh the data resident on the devices between training calls.

\paragraph{Algorithms.}
An algorithm in \boonta{} consists of three pure functions.
The first creates the algorithm state, such as parameters and optimizer state, from an initial timestep, the second chooses actions for a batch of timesteps, and the third turns a batch of transitions into a new algorithm state.
The podracer calls the second once per environment step and the third once per batch, so an algorithm never sees the training loop.
The algorithm receives its networks and optimizers. DQN uses a network over discrete action values. SAC maintains an actor and critics for continuous actions. Recurrent GRPO can use a pretrained transformer to generate actions one token at a time.
The policy is the network itself, whose output head returns a distribution, and evaluation chooses actions with the same function at temperature zero, which makes every head greedy.
The recurrent state is kept out of the algorithm state.
Like the environment state, it belongs to the rollout, so the podracer keeps it and records it at every step.
Checkpoints then hold only what was learned, and a recurrent update can replay a rollout from the state it started with.

\paragraph{Environments.}
Environments create and advance their state, and they can change the parameters of a running state, such as the current level, which is how curricula reach them.
The same interface covers environments written in JAX, Gymnasium vector environments \citep{towers2024gymnasium}, and native simulators such as the Arcade Learning Environment \citep{bellemare2013ale}, which are called through the JAX foreign function interface.
Each step returns a timestep that distinguishes termination from truncation, so that value targets bootstrap through a truncation but not through a termination.
Episodes reset in the same step in which they end, so the timestep that ends one episode already holds the first observation of the next, and its termination flag also tells recurrent networks where to reset their state.
Automatic resets, vectorization, time limits, reward transformations and fixed opponents are environment wrappers, and podracers interact only with batched environments.

\section{Distributed training}
\label{sec:distributed}

\paragraph{Device placement.}
A data-parallel learner needs different placements for different parts of its state. For example, DQN keeps a copy of its online and target network parameters on each device, while its replay buffer grows with the number of environments.
An algorithm marks the fields that hold one entry per environment with the data axis in the field's metadata.
The podracer derives the placement of the whole training state from these marks.
Marked fields are partitioned along the data axis of the device mesh, while all other fields, including parameters and optimizer state, are replicated.
Environment states, timesteps and recurrent states are always partitioned, and the compiler inserts the communication between devices that the computation requires.
The number of environments must be divisible by the number of devices.
For several hosts, \texttt{hangar} starts one process per GPU, for example with \texttt{srun --nodes=2 --ntasks-per-node=4 boonta train num\_processes=8}, initializes \texttt{jax.distributed}, and builds the mesh over the devices of all processes.
Anakin and Quadinaros use this mesh across hosts. Sebulba currently runs on a single host.

\paragraph{Rollout minibatches.}
PQN computes action-value targets and PPO computes advantages and returns from the latest rollout. Both reuse these quantities across update epochs. Both learners permute indices and gather the current minibatch, avoiding a second, fully shuffled copy of the rollout.
With $m$ minibatches, each learner groups adjacent entries by environment into blocks of size $m$ and permutes the indices within each block. Minibatch $j$ takes the $j$th entry of each permutation. Each transition is used once per epoch, and a device holding complete blocks can gather its observations locally. The devices combine gradients to update the replicated parameters.

\paragraph{Replay sampling.}
DQN and SAC add fresh transitions to a replay buffer and repeatedly draw batches from it. The buffer is partitioned by environment so each device can write the trajectories it collected. The sampler groups environment rows into equal-sized blocks and draws an equal share of the training batch from each block. With blocks local to devices, each learner replica reads its portion of the batch locally. The buffer retains old transitions across updates, so increasing the environment count at fixed history per environment also increases total replay capacity.
\section{Policies and datasets}

\paragraph{Networks.}
A network consists of a feature extractor, an optional torso and a head.
Every torso is a block that receives a recurrent state, an input and the episode-start flags and returns a new recurrent state and an output, so recurrent cells evaluated with a sequential scan, recurrences evaluated with a parallel scan, and self-attention with a key-value cache can be exchanged for one another.
Pretrained transformers such as Qwen3 \citep{qwen2025qwen3} are implemented as such blocks, with the key-value cache as the recurrent state, and LoRA \citep{hu2022lora} can wrap any block, so the algorithm only operates on the trainable parameters.

\paragraph{Mixed precision and recurrent kernels.}
The network configures the compute and parameter dtypes separately, so a policy can use bfloat16 (BF16) matrix multiplies with FP32 parameters. The state-space layers also separate the precision of the recurrent accumulation from the precision of its inputs and outputs.
For an elementwise recurrence $h_t = a_t h_{t-1} + b_t$, \boonta{} provides a Pallas kernel and a custom backward pass. The kernel processes independent batch elements and groups of features in parallel, iterates over time within each group, and accumulates the carry in FP32. The default implementation selects this kernel on NVIDIA GPUs of compute capability 9.0 or higher for feature widths divisible by 128 and otherwise uses a JAX associative scan.

\paragraph{Pretrained models.}
A pretrained transformer is a network whose recurrent state is its key-value cache, so a language model can be trained as a recurrent policy without a separate training stack, with every generated token as one action.
The recipes for games played through tool calls post-train the instruction-tuned Qwen3-0.6B and Qwen3-4B \citep{qwen2025qwen3} with recurrent GRPO, using rank-8 LoRA adapters \citep{hu2022lora} on quantized weights and bfloat16 computation.
The \texttt{RecurrentGRPO} implementation it uses contains no code specific to language, which lives in the recipe, the environment wrappers and the network.
\boonta{} also loads the vision-language-action model $\pi_0$ \citep{black2024pi0} with its Gemma and SigLIP components and provides the LIBERO benchmark \citep{liu2023libero} as an environment.

\paragraph{Datasets.}
For BC and offline RL, the size of the dataset and the size of a learner batch can be chosen independently. The Minari adapter can keep the full dataset resident or sample a pool of transitions from episodes on the host. In the latter case a background thread prepares the next pool while Quadinaros trains on the current one. The pool is partitioned along the data axis and the learner samples minibatches on device. For recurrent BC, the Kinetix adapter reads trajectory batches from shards and renders observations from the recorded environment states.

\section{Software}
\label{sec:software}

\paragraph{Installation.}
\boonta{} is released under the Apache 2.0 license and is installed from source with uv.
The core library depends on JAX, Flax, Optax, Distrax, Flashbax, Orbax and lox, which is installed from GitHub, while environment suites, Hugging Face model loading and the experiment tooling are optional extras.
The algorithms, datasets, environments, networks and podracers packages each have a guide that documents the interface and describes how to add a new component, and a quickstart notebook covers training and evaluation in plain Python.

\paragraph{hangar.}
The distribution also contains \texttt{hangar}, an experiment framework built on the library and similar in role to RL Baselines3 Zoo for Stable-Baselines3 \citep{raffin2021sb3, raffin2020zoo}.
\texttt{hangar} imports \boonta{}, while no module of the library imports \texttt{hangar}, Hydra or OmegaConf.
\texttt{hangar} provides the \texttt{boonta} command, whose \texttt{train} and \texttt{evaluate} subcommands build a podracer from a Hydra configuration \citep{yadan2019hydra}, run it, and handle checkpointing, resumption and multi-host execution, for example \texttt{boonta train algorithm=ppo environment=gymnax/minatar/breakout}.
It contains 43 recipes, one per combination of algorithm and environment suite, which cover 57 registered configurations.
Hyperparameters for these configurations are stored in 67 files, and Appendix~\ref{app:reproduction} lists those of every configuration in this paper. Where a configuration follows a published setting, the appendix names its source, and the remaining settings are the defaults of \texttt{hangar}. Hydra plugins add CARBS \citep{fetterman2023carbs} and population-based training \citep{jaderberg2017pbt} sweepers as well as a launcher for Determined clusters.

\paragraph{Logging and artisans.}
None of the interfaces return metrics.
Algorithms, wrappers and pits record values with the lox library \citep{lox} at the point where they are computed, inside compiled code.
When a podracer is constructed, lox collects the values logged during each training or evaluation call and returns them alongside the state, keeping training and evaluation metrics apart.
Adding a metric to an algorithm therefore changes no interface.
Other outputs of a run, such as checkpoints, videos and TrueSkill ratings, are produced by \emph{artisans}, which observe the podracer state outside the training loop, and loggers write them to disk, to a terminal dashboard or to Weights \& Biases.

\paragraph{Implemented methods.}
\boonta{} includes PPO \citep{schulman2017ppo}, REPPO \citep{voelcker2025reppo}, PQN \citep{gallici2024pqn}, GRPO \citep{shao2024deepseekmath} and MMD \citep{sokota2023mmd}, which learn from the current rollout, the off-policy algorithms DQN \citep{mnih2015dqn} and SAC \citep{haarnoja2018sac}, and the offline algorithms BC and IQL \citep{kostrikov2022iql}.
The evolution strategy EGGROLL \citep{sarkar2025eggroll} implements the same three functions. Each update is one generation, which scores mirrored low-rank perturbations of the policy weights on the rollouts of the batch.
Most algorithms have a recurrent variant, whose sequence model may be a recurrent cell, a recurrence evaluated with a parallel scan, or self-attention, and most accept a tuple of auxiliary losses that are added to the training objective.
Multi-agent methods need no separate algorithm.
IPPO \citep{dewitt2020ippo} and MAPPO \citep{yu2022mappo} are the PPO implementation with a separate critic for each agent and with a critic over the observations of all agents, respectively, configured in their recipes.
Prioritized level replay \citep{jiang2021prioritized} and a league with self-play, fictitious self-play, prioritized fictitious self-play and PSRO \citep{lanctot2017unified} as meta-solvers are implemented as curricula, without modifications to the algorithms they are used with.
PLR has configurations for PPO on Kinetix and MinAtar and for recurrent PPO on the JaxUED maze and CartPole, and the experiments train it with both Anakin and Sebulba. The league has no configuration in \texttt{hangar} yet and is exercised by tests with Anakin.

\paragraph{Curricula, pits and laps.}
Many methods that adapt the training distribution operate between updates.
Prioritized level replay scores levels using the agent's rollouts and selects the levels to replay \citep{jiang2021prioritized}, and population-based methods record match outcomes and select opponents with a meta-solver \citep{lanctot2017unified}.
Such methods are commonly implemented by modifying the training loop of a specific algorithm, as in the JaxUED scripts discussed in Appendix~\ref{app:design}.
\boonta{} instead provides two hooks that every podracer invokes, the \emph{pit} between updates and the \emph{lap} after every environment step.
A pit has the same structure as an algorithm. It keeps its own state, updates that state from the transitions of each batch, and then acts, except that its actions change the environment, for example by assigning levels, rather than choosing the agent's moves.
After each algorithm update, the pit learns from the transitions of that update without touching the environment, and when it acts, it returns the podracer state with new levels assigned or new opponents seated.
In population-based training, it may also change the algorithm state, for instance to reinitialize a newly admitted learner.
The pit acts at initialization, after every update, and after evaluation resets the environment, so that evaluation uses the same environment configuration as training.
Pits are pure functions, and their state is part of the podracer state, so they are compiled into the training program and placed on devices with the rest of the state.
The lap is a pit without the learning step, meant for state that the environment must see at every step.
It acts after every environment step, and at initialization and evaluation just before the pit, and since it can read the latest timestep and change its own state in the same call, it needs no separate update.
A lap that keeps one value per environment marks that field with the data axis, so that its state is split across devices with the environments.
No curriculum in the library currently uses the lap.
A curriculum is a function that takes the algorithm and the environment and returns possibly wrapped versions of both, together with an optional pit and lap.
Prioritized level replay, for example, wraps the environment so that its levels can be set and adds a pit that maintains the level buffer.
The league wraps the algorithm in a population of learners and the environment in an opponent wrapper, and its pit selects matchups with a meta-solver such as fictitious self-play or PSRO.
Sebulba does not support populations, so the league runs only on Anakin.
Since curricula are applied at a fixed position in the environment stack, after automatic resets and before episode statistics are recorded, one curriculum can be replaced by another without changing the surrounding code.

\section{Experiments}
\label{sec:experiments}

The experiments follow the claims of the abstract. We measure how throughput scales across the GPUs of one node, with a growing and with a fixed global batch, compare synchronous and asynchronous training of the same PPO implementation, and measure the effect of reduced precision. We then train the same algorithm implementations across the range of settings, from CartPole to the post-training of a language model, and evaluate online learners on games and continuous control tasks and offline learners on fixed expert datasets.

Runs are launched through \texttt{hangar} from configurations in the repository, and Appendix~\ref{app:commands} gives the command for each.
Unless stated otherwise, we run 10 seeds per configuration and report the mean with 95\% bootstrap confidence intervals for each task, and the interquartile mean with stratified bootstrap confidence intervals when aggregating over tasks \citep{agarwal2021precipice}.
The reported training runs use NVIDIA A100 GPUs with 80\,GB of memory, eight to a node. Within a node, every pair of GPUs is connected by 12 NVLink links through NVSwitch, and each node has four network adapters.

\begin{figure}[t]
  \centering
  \includegraphics[width=\linewidth]{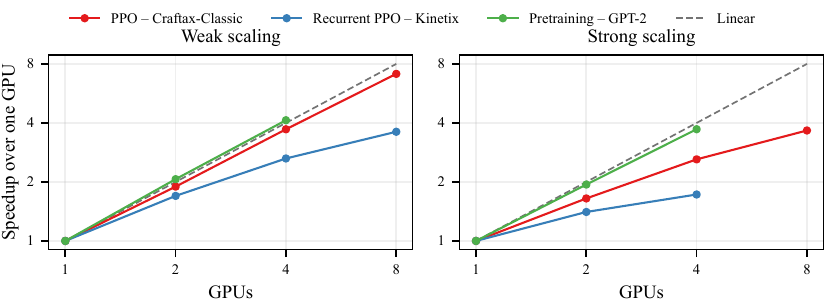}
  \caption{Speedup in throughput over one GPU with Anakin on a node of eight A100 GPUs, in environment steps per second for PPO and recurrent PPO and in tokens per second for GPT-2. Left, weak scaling with a fixed batch per GPU. Right, strong scaling with a fixed global batch, averaged over three seeds for Craftax-Classic and from one seed otherwise. The dashed line marks linear scaling.}
  \label{fig:scaling}
\end{figure}

\paragraph{Multi-GPU scaling.}
We train three workloads of different cost per step with Anakin, namely PPO on Craftax-Classic \citep{matthews2024craftax}, recurrent PPO with a GRU on Kinetix \citep{matthews2025kinetix}, whose environments simulate rigid-body physics, and recurrent behavior cloning that pretrains GPT-2 on FineWeb-Edu. The first two run on up to eight GPUs of one node and GPT-2 on up to four.
For weak scaling, each device keeps the same batch, 1024 environments on Craftax-Classic, 8192 on Kinetix and 32 sequences of 1024 tokens for GPT-2, so the global batch grows with the number of devices.
With eight GPUs, PPO on Craftax-Classic processes 7.1 times as many environment steps per second as on one GPU, an efficiency of 89\%, and recurrent PPO on Kinetix 3.6 times. GPT-2 scales almost linearly, to 4.1 times the tokens per second on four GPUs (Figure~\ref{fig:scaling}, left).
Kinetix scales least, since its recurrent update over 64-step sequences is the largest share of its step and its environments reset through a fixed pool of fresh levels per step.
PPO on Breakout-MinAtar and DQN on Breakout-MinAtar reach 4.6 and 6.5 times on eight GPUs. Breakout scales less than Craftax-Classic because its small network leaves little computation per environment step to spread across devices.
For strong scaling, the global batch stays fixed, at 32,768 environments on Craftax-Classic and Kinetix and 64 sequences for GPT-2, so runs use the same batch and the same minibatches per update.
At this fixed workload, eight GPUs deliver $3.7\times$ the single-GPU throughput on Craftax-Classic, and four GPUs deliver $3.7\times$ for GPT-2 and $1.7\times$ on Kinetix, which we measured up to four GPUs (Figure~\ref{fig:scaling}, right). The 996M-step Craftax run takes an estimated 4.6 minutes on eight GPUs and 16 minutes on one.
We estimate elapsed training time by summing the steps in each logged interval divided by its measured throughput. The mean time to the first logged training return of at least 10 on Craftax-Classic is 205, 133, 92 and 72 seconds on 1, 2, 4 and 8 GPUs, respectively, over three seeds per configuration, and the final return is 17.0 to 17.1 with every number of devices.

\paragraph{Asynchronous training.}
We train the same PPO implementation on Breakout-MinAtar with the same JAX environment under Anakin and Sebulba (Appendix~\ref{app:example}).
Anakin trains on one GPU, and Sebulba on two, one acting and one learning.
Figure~\ref{fig:podracers} in Appendix~\ref{app:reproduction} shows greedy evaluations of the saved checkpoints, including the untrained policies at step~0.
Anakin has the higher mean return throughout training, and the confidence intervals overlap. After 19.7M steps, the mean return is 69.1 [53.6, 82.8] with Anakin and 61.9 [49.3, 73.4] with Sebulba, with 95\% bootstrap confidence intervals.
The mean of the per-run median throughputs is 3.72M environment steps per second for Anakin and 3.53M for Sebulba.
These rates describe the two complete configurations. Both the device count and the collection schedule differ.

\paragraph{Precision and kernels.}
We train recurrent PPO with a MinGRU torso on Breakout-MinAtar on one A100 with FP32 and with BF16 computation in the hidden layers, with three seeds each. Only \texttt{network.dtype} changes, and the policy logits and the value stay in FP32.
BF16 raises throughput from 2.30M to 3.12M environment steps per second, by 36\%, and the final returns are 6.5 with FP32 and 6.7 with BF16.
The Pallas kernel for the recurrence requires a GPU of compute capability 9.0, and the A100 has 8.0, so these runs use the associative scan.

\paragraph{From CartPole to language models.}
PPO trains on Gymnasium CartPole with Sebulba, where the environments step in CPU processes while one GPU acts and another learns. Over 10 seeds, 7 runs end at the maximum return of 500 and one at 498, the mean final return is 426 [316, 500], and training runs at about 21.3K environment steps per second.
Pretraining a language model is behavior cloning of a recurrent policy whose actions are tokens and whose recurrent state is the KV cache. Recurrent BC trains GPT-2 with 124M parameters \citep{radford2019language} on 123M tokens of FineWeb-Edu \citep{penedo2024fineweb} in BF16 on one A100. Over three seeds, the negative log-likelihood of the next token falls from 8.5 to 3.88 at 59.5K tokens per second.
Recorded interactions of an agent mix its own decisions with the observations it received, such as user requests and tool output, and behavior cloning in the POMDP view predicts only the decisions. With the same recurrent BC implementation, we train Qwen3-style models of 82M, 191M and 504M parameters from scratch on the agent interactions of AgentTrove \citep{openthoughts2026agenttrove}, either on every token or only on the assistant's tokens, while both read the whole interaction. On held-out assistant tokens, training only on the assistant's tokens reaches a negative log-likelihood of 1.35 and 1.12 with 82M and 191M parameters, against 1.47 and 1.19 when training on every token, and 1.34 against 1.33 with 504M (Table~\ref{tab:blm} in Appendix~\ref{app:reproduction}). These are single runs at matched estimated compute. The estimate charges the output layer only at scored positions, so the runs on assistant tokens see more training positions.
Post-training is RL of the same kind of policy. Recurrent GRPO post-trains Qwen3-0.6B \citep{qwen2025qwen3} with LoRA \citep{hu2022lora} on quantized weights to find a secret whole number from 1 to 30 in at most five guesses. Each guess is a call to a tool that replies whether the secret is higher or lower. The environment gives the prompt as its first observation, each guess is written one token at a time, the reply of the tool is the next observation, and a win earns 1. Over three seeds, the share of games won rises from 10\% to 31\% in 2.1 hours on one A100 at about 110 tokens per second, with final win rates of 23\%, 29\% and 42\% (Figure~\ref{fig:post-training} in Appendix~\ref{app:reproduction}). Binary search wins every game, so the model remains far from the best policy. Wordle, played through the same kind of tool, is harder. Qwen3-4B, which also earns a small partial credit for each correct letter, raises its partial credit in two of three seeds within 2M tokens but wins no game, and the third seed collapses to a return of zero.
All three use implementations unchanged from the other experiments.

\paragraph{Throughput.}
Moving the training loop out of the algorithm could add overhead, so we compare \boonta{} with Stoix \citep{toledo2024stoix}, which also implements the Podracer architectures but writes the training loop into each algorithm.
We run PPO with Anakin in both libraries on CartPole, Breakout-MinAtar, Craftax-Classic and Kinetix, with the same hyperparameters, network architectures and environment implementations, three seeds and one A100.
We vary the number of environments, since a fixed cost per step is most visible at small batch sizes, and Figure~\ref{fig:throughput} shows the environment steps per second relative to Stoix. Table~\ref{tab:throughput} in Appendix~\ref{app:reproduction} also gives the final returns.
With default settings, \boonta{} reaches 101\% to 105\% of Stoix's throughput on Breakout-MinAtar and 121\% and 142\% on CartPole with 8192 and 65536 environments. With 64 and 1024 environments, CartPole reaches 89\% and 90\%. There, the environment and the network cost almost nothing, so the time of an update is set by the number of small kernels it launches.
Computing the hidden layers in BF16 and resetting environments optimistically changes no part of the algorithm and raises throughput to 1.21 and 1.49 times Stoix's on Breakout-MinAtar with 8192 and 65536 environments and to 1.14 and 1.74 times on CartPole, while it lowers throughput with small batches, where the extra casts and resets are not amortized. Optimistic resets in Stoix show the same drop with 1024 environments, to 46\% on CartPole and 63\% on Breakout-MinAtar, and do not raise its throughput on these two environments with more environments.
The final returns are similar on Breakout-MinAtar and on CartPole with 64 and 65536 environments. With 1024 and 8192 environments, Stoix ends with higher returns on CartPole, 400 and 289 against 280 and 236, and it differs from \boonta{} in optimizing the actor and the critic with separate optimizers.
On Craftax-Classic, \boonta{} with default settings reaches 74\% to 84\% of Stoix's throughput, and BF16 with optimistic resets raises it to 1.32 to 2.37 times Stoix's, while optimistic resets alone raise Stoix's own throughput by 10\% to 42\%. The final returns are similar.
On Kinetix, where generating a random level makes every reset expensive, both libraries reach the same throughput with default settings, and optimistic resets raise it 5 to 22 times in both (Table~\ref{tab:throughput}). The returns stay near zero within the 2M to 20M steps of these runs, so they show throughput only.
Sebulba PPO on Atari Breakout with one actor GPU and one learner GPU reaches 1.43 and 1.34 times Stoix's throughput with 128 and 512 environments. Stoix steps the game with EnvPool and \boonta{} with the vectorized environment of Gymnasium, so this comparison includes the simulators, and we do not compare its returns.
Appendix~\ref{app:overhead} compares the auto-reset strategies.

\begin{figure}[t]
  \centering
  \includegraphics[width=\linewidth]{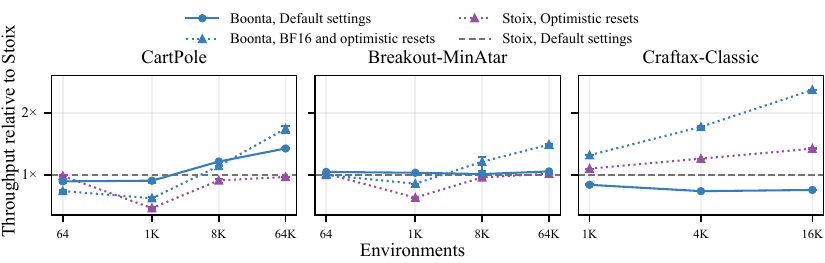}
  \caption{Environment steps per second of PPO with Anakin on one A100 in \boonta{} and in Stoix with optimistic resets, relative to Stoix with default settings, all with the same hyperparameters, network architectures and environments. Default settings use same-step resets and FP32. \boonta{} also runs with BF16 hidden layers and optimistic resets. Points show the mean over three seeds divided by the mean of Stoix with default settings, and bars the range over the seeds. Table~\ref{tab:throughput} gives the absolute numbers.}
  \label{fig:throughput}
\end{figure}

\paragraph{Curricula.}
We train recurrent PPO on JaxUED mazes \citep{coward2024jaxued} with domain randomization (DR) and prioritized level replay (PLR) \citep{jiang2021prioritized} on Anakin, and with PLR on Sebulba.
We evaluate the final checkpoints greedily on the same eight held-out mazes (Figure~\ref{fig:curricula} in Appendix~\ref{app:reproduction}).
Averaged equally across the eight mazes, the mean returns are 0.21 for DR on Anakin, 0.16 for PLR on Anakin, and 0.13 for PLR on Sebulba.

\paragraph{Online and offline benchmarks.}
For online RL, we train PPO, DQN and PQN on four MinAtar games \citep{young2019minatar} and SAC on five Brax versions of the MuJoCo tasks.
Figure~\ref{fig:minatar} compares Q-learning from replay (DQN), Q-learning from recent rollouts (PQN) and policy-gradient updates (PPO) in the same environments. Figure~\ref{fig:sac} gives SAC learning curves across the five continuous control tasks.
For offline RL, we train BC and IQL with Quadinaros on three Minari expert datasets \citep{younis2024minari}. Played greedily in the Gymnasium tasks the datasets were recorded in, the five final checkpoints of BC score 3616, 6906 and 16868 on Hopper, Walker2d and HalfCheetah, and those of IQL 3664, 6883 and 15579 (Figure~\ref{fig:offline} in Appendix~\ref{app:reproduction}). Relative to the mean return of the expert data, BC reaches 94\%, 101\% and 104\%, and IQL 95\%, 101\% and 96\%.
Multi-agent RL is not part of this comparison, since the published SMAX results of JaxMARL \citep{rutherford2024jaxmarl} use recurrent policies and the IPPO and MAPPO configurations of \texttt{hangar} are feed-forward.
The configured runs use the settings provided by \texttt{hangar}, with the quickstart notebook settings for PPO on Breakout.
We also train the same DQN implementation for 5M steps on official MinAtar and on gymnax with the same configuration wherever supported (Figure~\ref{fig:dqn-minatar}).
Appendix~\ref{app:reproduction} gives training curves and describes differences from the published settings.
The Minari datasets contain expert data only, so the D4RL results of the IQL paper do not apply directly.

\section{Conclusion}
\label{sec:conclusion}

\boonta{} separates what an RL agent learns from how its training is scaled.
Algorithms consist of three pure functions and mark the state that belongs to individual environments, while podracers, networks, buffers and datasets provide placement across devices and hosts, asynchronous collection, reduced precision, custom kernels and data staging.
On one node of eight A100 GPUs, the unchanged algorithms reach 4.6 to 7.1 times the single-GPU throughput for PPO and 6.5 times for DQN with a fixed number of environments per GPU.
Future work includes measuring multi-node throughput, extending Sebulba across hosts and partitioning parameters for larger policies.

\paragraph{Limitations.}
The episode buffers sample over the whole buffer, so their update is still computed in full on every device.
The device mesh partitions data and replicates parameters. Each device must therefore hold a complete copy of the policy weights, including the frozen weights when training adapters.
Sebulba is restricted to a single host and does not support the league.
The Gymnasium adapter transfers data to the host at every step, which is acceptable for simulators whose step time is on the order of milliseconds but limits throughput for faster environments.
\boonta{} and its logging dependency lox are installed from source, and a release on PyPI is planned.

\bibliographystyle{plainnat}
\bibliography{references}

\appendix

\section{Commands}
\label{app:commands}

Each experiment is a \texttt{boonta train} command, and every command takes \texttt{seed=}$n$. The PPO runs on Breakout-MinAtar add the settings of the quickstart notebook in Table~\ref{tab:hyperparameters}, the throughput runs add the settings listed there for them, and \texttt{boonta evaluate} with the same arguments and \texttt{checkpoint=} plays a saved run greedily.

\begin{lstlisting}[language=bash]
# Multi-GPU scaling, weak and strong
boonta train algorithm=ppo environment=craftax/craftax_classic/symbolic \
    podracer.config.mesh.count=8 environment.num_envs=8192
boonta train algorithm=ppo environment=craftax/craftax_classic/symbolic \
    podracer.config.mesh.count=8 environment.num_envs=32768 num_steps=1_000_000_000
# Anakin and Sebulba
boonta train algorithm=ppo environment=gymnax/minatar/breakout
boonta train algorithm=ppo environment=gymnax/minatar/breakout \
    podracer=sebulba podracer.config.actor.start=1
# Throughput, with 64 to 65536 environments
boonta train algorithm=ppo environment=gymnax/classic_control/cartpole \
    environment.num_envs=8192 num_steps=50_000_000 \
    ++network.dtype=bfloat16 environment/auto_reset=optimistic
# Precision
boonta train algorithm=recurrent_ppo environment=gymnax/minatar/breakout \
    cell=min_gru environment.num_envs=8192 podracer.config.num_steps=64 \
    ++network.dtype=bfloat16
# From CartPole to language models
boonta train algorithm=ppo environment=gymnasium/cartpole \
    podracer=sebulba podracer.config.actor.start=1
boonta train algorithm=recurrent_bc environment=fineweb
boonta train algorithm=recurrent_grpo environment=dune_sea/guess_number \
    num_steps=1_000_000
boonta train algorithm=recurrent_grpo environment=dune_sea/wordle \
    network.repo_id=Qwen/Qwen3-4B optimizer.lr=5e-5 environment.num_envs=16 \
    num_steps=2_000_000
# Curricula
boonta train algorithm=recurrent_ppo environment=jaxued/maze curriculum=plr
# Benchmarks
boonta train algorithm=dqn environment=gymnax/minatar/breakout
boonta train algorithm=pqn environment=gymnax/minatar/breakout
boonta train algorithm=sac environment=brax/mujoco/hopper
boonta train algorithm=iql environment=brax/mujoco/hopper
boonta evaluate algorithm=iql environment=gymnasium/mujoco/hopper \
    checkpoint=<run directory>
\end{lstlisting}

\section{Example}
\label{app:example}

The following condensed example from the quickstart notebook trains PPO on MinAtar Breakout with Anakin, and then with Sebulba with actors and learner on separate devices.
Only the podracer changes between the two, while the environment, the network and the algorithm stay the same objects.
Hyperparameters are omitted.

\begin{lstlisting}
env = environments.make(namespace="gymnax", env_id="Breakout-MinAtar")
env = SameStepAutoReset(env, num_envs=8192)
env = RecordEpisodeStatistics(env, gamma=0.989)

network = Network(
    feature_extractor=FeatureExtractor(observation_extractor=cnn),
    head=ActorCritic(actor=Categorical(nn.Dense(num_actions)), critic=nn.Dense(1)),
)
algorithm = PPO(cfg=PPOConfig(...), network=network, optimizer=optax.adam(2.4e-3))

podracer = anakin.make(
    config=anakin.AnakinConfig(num_steps=16, mesh=mesh(jax.device_count())),
    algorithm=algorithm,
    environment=env,
)
podracer = sebulba.make(
    config=sebulba.SebulbaConfig(
        num_steps=16, learner=mesh(1), actor=mesh(1, start=1)
    ),
    algorithm=algorithm,
    environment=env,
)

state = podracer.init(jax.random.key(0))
state, logs = podracer.train(state, jax.random.key(1), num_updates=75)
\end{lstlisting}

\section{Environments}
\label{app:environments}

Table~\ref{tab:environments} lists the environment suites \boonta{} supports. Environments written in JAX are compiled into the training program. Python environments run through a JAX callback, and native emulators are called through the JAX foreign function interface, so that the podracers drive all of them in the same way.

\begin{table}[h]
  \centering
  \caption{Environment suites and how each is integrated.}
  \label{tab:environments}
  \resizebox{\linewidth}{!}{%
  \begin{tabular}{lll}
    \toprule
    Integration & Suite & Domain \\
    \midrule
    In JAX & gymnax & Classic control, bsuite, MinAtar \\
     & Brax, MuJoCo Playground & Continuous control, with JAX or Warp physics \\
     & Craftax, XLand-MiniGrid & Open-ended games \\
     & Kinetix, JaxUED & Physics and maze tasks for curricula \\
     & Jumanji, JAXtari & Combinatorial problems and Atari games \\
     & JaxMARL, Mapox, BluffJAX & Multi-agent and imperfect-information games \\
     & Dune sea & T-maze, chain of forks, ConnectX, Wordle and number guessing \\
    \midrule
    Callback & Gymnasium & Any vector environment, including MinAtar and MuJoCo \\
     & Isaac Lab, Isaac Lab Arena & GPU robotics simulation \\
     & LIBERO & Robot manipulation for vision-language-action models \\
    \midrule
    Foreign function interface & ALE & Atari 2600 \\
     & Peanut-GB & Game Boy, including Pok\'emon Red \\
     & pyNES & NES, including Super Mario Bros. \\
    \bottomrule
  \end{tabular}
  }
\end{table}

\section{Auto-reset strategies}
\label{app:overhead}

Table~\ref{tab:autoreset} compares the three auto-reset strategies with PPO on Breakout-MinAtar, using the settings of the quickstart notebook for 20M steps and 10 seeds on one GPU.
All three train at 4.99M to 5.16M environment steps per second, and their final returns do not differ significantly (Welch's $t$-test, $p \geq 0.15$ for every pair).

\begin{table}[h]
  \centering
  \caption{Throughput and final return of PPO on Breakout-MinAtar with each auto-reset strategy, with 95\% bootstrap confidence intervals over 10 seeds.}
  \label{tab:autoreset}
  \begin{tabular}{lrr}
    \toprule
    Auto-reset & Steps per second & Final return \\
    \midrule
    Same step & 5.16M & 64.9 [58.6, 71.0] \\
    Next step & 4.99M & 59.7 [49.4, 68.2] \\
    Optimistic & 5.03M & 52.5 [36.8, 65.0] \\
    \bottomrule
  \end{tabular}
\end{table}

\section{Reproduction details}
\label{app:reproduction}

Tables~\ref{tab:hyperparameters} and~\ref{tab:hyperparameters-other} give the settings of every configuration in this paper as composed by \texttt{hangar}. The PPO runs of the throughput comparison on CartPole and Breakout-MinAtar use the PureJaxRL settings, the DQN runs on official MinAtar use the settings of the MinAtar paper \citep{young2019minatar}, and IQL scales its rewards as its paper does \citep{kostrikov2022iql}, with a temperature of 1 on HalfCheetah.
Stoix runs its Anakin PPO at commit \texttt{8fff19c} with the same hyperparameters, environments, layer widths and activations as \boonta{}, and its Sebulba PPO on Atari Breakout with the Nature network in both libraries. Its optimistic resets prepare one reset for every 64 environments, as in \boonta{}. Stoix excludes evaluation from its reported steps per second, and we report that number.

\begin{table}[h]
  \centering
  \small
  \caption{Settings of the on-policy configurations. Environments and rollout length give the batch of one update.}
  \label{tab:hyperparameters}
  \setlength{\tabcolsep}{4pt}
  \resizebox{\linewidth}{!}{%
  \begin{tabular}{lrrrrrrrrr}
    \toprule
    Configuration & Envs & Rollout & Minibatches & Epochs & LR & $\gamma$ & $\lambda$ & Clip & Entropy \\
    \midrule
    PPO -- Breakout-MinAtar (quickstart) & 8192 & 16 & 16 & 4 & $2.4 \times 10^{-3}$ & 0.989 & 0.77 & 0.14 & 0.003 \\
    PPO -- Other MinAtar & 4096 & 16 & 16 & 2 & $2.5 \times 10^{-4}$ & 0.99 & 0.95 & 0.2 & 0.01 \\
    PPO -- Craftax-Classic & 32768 & 16 & 16 & 2 & $2.5 \times 10^{-4}$ & 0.99 & 0.95 & 0.2 & 0.01 \\
    PPO -- Gymnasium CartPole & 16 & 128 & 4 & 4 & $3 \times 10^{-3}$ & 0.99 & 0.95 & 0.2 & 0.01 \\
    PPO -- CartPole (throughput) & 64--65536 & 128 & 4 & 4 & $2.5 \times 10^{-4}$ & 0.99 & 0.95 & 0.2 & 0.01 \\
    PPO -- Breakout-MinAtar (throughput) & 64--65536 & 128 & 8 & 4 & $5 \times 10^{-3}$ & 0.99 & 0.95 & 0.2 & 0.01 \\
    PPO -- Craftax-Classic (throughput) & 1024--16384 & 64 & 8 & 4 & $2 \times 10^{-4}$ & 0.99 & 0.8 & 0.2 & 0.01 \\
    PPO -- Kinetix (throughput) & 1024--16384 & 64 & 8 & 4 & $3 \times 10^{-4}$ & 0.99 & 0.95 & 0.2 & 0.01 \\
    PPO -- Atari Breakout (throughput) & 128--512 & 128 & 4 & 4 & $2.5 \times 10^{-4}$ & 0.99 & 0.95 & 0.1 & 0.01 \\
    PQN -- MinAtar & 4096 & 16 & 16 & 1 & $5 \times 10^{-4}$ & 0.99 & 0.65 & -- & -- \\
    Recurrent PPO -- Mazes & 32 & 256 & 1 & 5 & $10^{-4}$ & 0.995 & 0.98 & 0.2 & 0.001 \\
    Recurrent PPO -- MinAtar & 8192 & 64 & 8 & 4 & $2.5 \times 10^{-4}$ & 0.99 & 0.95 & 0.2 & 0.01 \\
    Recurrent GRPO -- Number guessing & 32 & 160 & 16 & 2 & $5 \times 10^{-5}$ & 1.0 & -- & 0.2 & -- \\
    Recurrent GRPO -- Wordle & 16 & 192 & 16 & 2 & $5 \times 10^{-5}$ & 1.0 & -- & 0.2 & -- \\
    \bottomrule
  \end{tabular}
  }
\end{table}

\begin{table}[h]
  \centering
  \small
  \caption{Settings of the off-policy, offline, pretraining and post-training configurations.}
  \label{tab:hyperparameters-other}
  \begin{tabular}{lp{0.6\linewidth}}
    \toprule
    Configuration & Settings \\
    \midrule
    DQN -- MinAtar & 128 environments, replay of 262{,}144, batch 512, learning rate $5 \times 10^{-4}$, $\tau = 0.005$, 10M steps \\
    SAC -- Brax & 128 environments, replay of $2^{20}$, batch 512, learning rates $6 \times 10^{-4}$, $\gamma = 0.997$, $\tau = 0.005$, 5M steps \\
    BC -- Minari & Batch 256, learning rate $3 \times 10^{-4}$, $10^6$ updates \\
    IQL -- Minari & Batch 256, learning rates $3 \times 10^{-4}$, expectile 0.7, temperature 3, $\tau = 0.005$, $10^6$ updates \\
    Recurrent BC -- GPT-2 & 12 layers of width 768, 6 sequences of 1024 tokens, learning rate $6 \times 10^{-4}$ with 2000 warmup updates and cosine decay to $6 \times 10^{-5}$, weight decay 0.1, 123M tokens \\
    Recurrent GRPO -- Number guessing & Qwen3-0.6B with LoRA of rank 8 on quantized weights, groups of 8, KL coefficient 0.04, 819K tokens \\
    Recurrent GRPO -- Wordle & Qwen3-4B with LoRA of rank 8 on quantized weights, groups of 8, KL coefficient 0.04, 2M tokens \\
    \bottomrule
  \end{tabular}
\end{table}

Figure~\ref{fig:minatar} shows the return of PPO, DQN and PQN on the gymnax versions of four MinAtar games with the settings that \texttt{hangar} provides.
Every checkpoint of a run, from the untrained policy at step 0, is played greedily with \texttt{boonta evaluate}.
PQN reaches a return of 58 on Freeway after 48M of its 80M steps and ends at 6.5, and DQN learns little on Asterix during its first 5M steps.

\begin{figure}[h]
  \centering
  \includegraphics[width=0.5\linewidth]{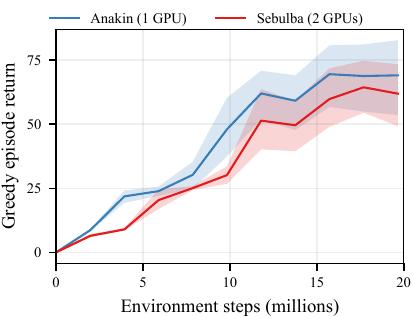}
  \caption{The same PPO implementation trained on Breakout-MinAtar with Anakin on one GPU and with Sebulba on two. Each point is the mean greedy return over 10 seeds. Shading shows 95\% bootstrap confidence intervals.}
  \label{fig:podracers}
\end{figure}

\begin{table}[h]
  \centering
  \caption{Environment steps per second and final return of PPO in Stoix and in \boonta{} with the same hyperparameters, network architectures, environments and hardware, with default settings and with optimistic resets, and in \boonta{} also with BF16 hidden layers, as means over three seeds. Atari Breakout runs Sebulba on two A100s, and there the optimized \boonta{} uses BF16 only. Its returns are left out because the two libraries use different simulators.}
  \label{tab:throughput}
  \resizebox{\linewidth}{!}{%
  \begin{tabular}{lrrrrrrr}
    \toprule
    & & \multicolumn{4}{c}{Steps per second} & \multicolumn{2}{c}{Final return} \\
    \cmidrule(lr){3-6} \cmidrule(lr){7-8}
    Environment & Environments & Stoix & Stoix optimistic & \boonta{} & \boonta{} optimized & Stoix & \boonta{} \\
    \midrule
    CartPole & 64 & 0.66M & 0.65M & 0.59M & 0.49M & 496 & 497 \\
    CartPole & 1024 & 7.75M & 3.57M & 7.01M & 4.81M & 400 & 280 \\
    CartPole & 8192 & 20.70M & 18.82M & 25.12M & 23.62M & 289 & 236 \\
    CartPole & 65536 & 27.43M & 26.45M & 39.07M & 47.86M & 141 & 133 \\
    Breakout-MinAtar & 64 & 0.41M & 0.42M & 0.43M & 0.41M & 66 & 64 \\
    Breakout-MinAtar & 1024 & 4.32M & 2.72M & 4.46M & 3.70M & 82 & 74 \\
    Breakout-MinAtar & 8192 & 10.13M & 9.70M & 10.24M & 12.26M & 34 & 28 \\
    Breakout-MinAtar & 65536 & 11.60M & 11.77M & 12.22M & 17.23M & 16 & 11 \\
    Craftax-Classic & 1024 & 0.17M & 0.19M & 0.14M & 0.22M & 11.6 & 12.0 \\
    Craftax-Classic & 4096 & 0.29M & 0.37M & 0.21M & 0.52M & 12.0 & 12.5 \\
    Craftax-Classic & 16384 & 0.33M & 0.47M & 0.25M & 0.79M & 9.8 & 10.8 \\
    Kinetix & 1024 & 4.3K & 21.3K & 4.4K & 21.9K & $-0.04$ & $-0.05$ \\
    Kinetix & 4096 & 4.8K & 60.7K & 4.9K & 61.0K & 0.00 & $-0.03$ \\
    Kinetix & 16384 & 4.9K & 105.9K & 4.9K & 104.1K & $-0.04$ & $-0.03$ \\
    Atari Breakout & 128 & 14.9K & -- & 21.3K & 21.8K & -- & -- \\
    Atari Breakout & 512 & 21.9K & -- & 29.3K & 31.8K & -- & -- \\
    \bottomrule
  \end{tabular}
  }
\end{table}

\begin{figure}[h]
  \centering
  \includegraphics[width=\linewidth]{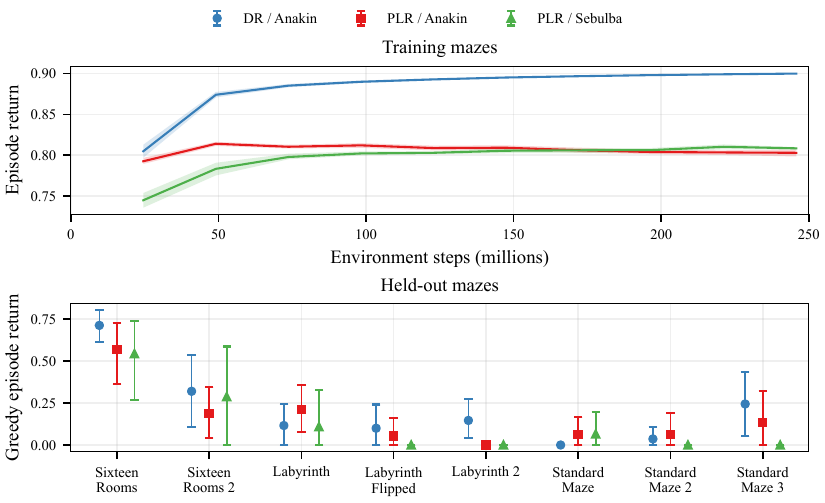}
  \caption{Recurrent PPO on JaxUED mazes. Top, returns on the mazes sampled by each curriculum during training. Bottom, greedy returns of the final checkpoints on eight fixed mazes. Means and 95\% bootstrap confidence intervals use 10 seeds for each Anakin configuration and five for Sebulba.}
  \label{fig:curricula}
\end{figure}

\begin{figure}[h]
  \centering
  \includegraphics[width=0.5\linewidth]{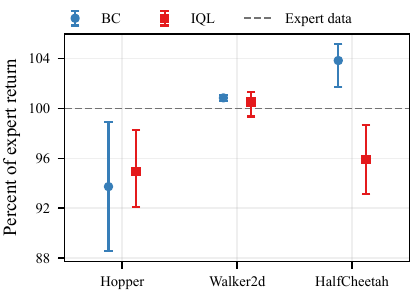}
  \caption{Greedy episode returns of the final BC and IQL checkpoints after training on Minari expert datasets, as a percentage of the mean return of the expert data, evaluated in the Gymnasium tasks the data were recorded in. Points and error bars show the mean and 95\% bootstrap confidence intervals over five seeds.}
  \label{fig:offline}
\end{figure}

\begin{figure}[h]
  \centering
  \includegraphics[width=0.5\linewidth]{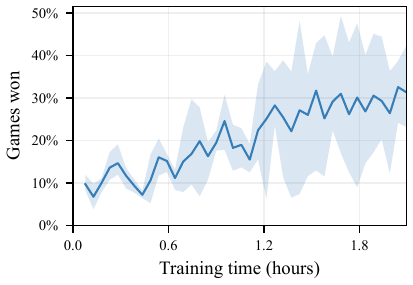}
  \caption{Share of number-guessing games won by Qwen3-0.6B during post-training with recurrent GRPO, with actions sampled at temperature 1, against the training time on one A100, estimated from the logged throughput. The line shows the mean and the band the 95\% bootstrap confidence interval over three seeds.}
  \label{fig:post-training}
\end{figure}

\begin{figure}[h]
  \centering
  \includegraphics[width=\linewidth]{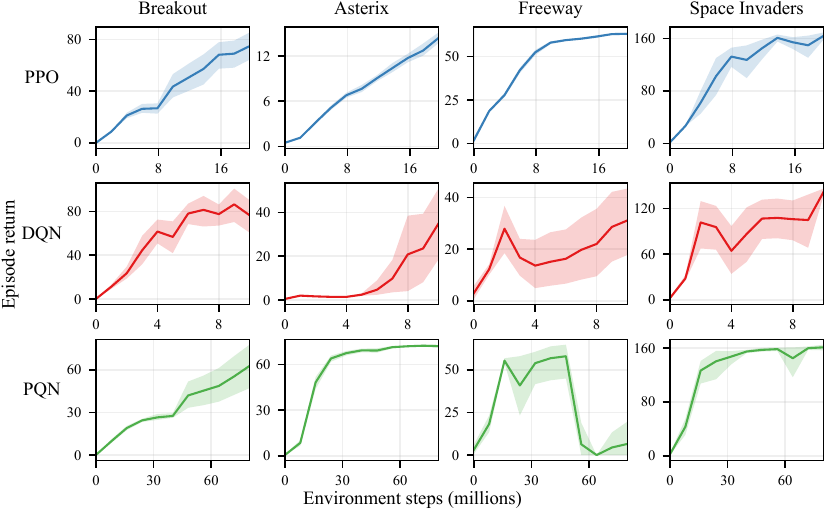}
  \caption{Greedy return of PPO, DQN and PQN checkpoints on four MinAtar games with the settings of \texttt{hangar}. Lines show the mean over 10 seeds and shaded regions 95\% bootstrap confidence intervals.}
  \label{fig:minatar}
\end{figure}

\paragraph{DQN on official MinAtar.}
The official environments use the full action set and a sticky-action probability of 0.1. We keep the DQN recipe's LayerNorm, Adam optimizer, squared TD loss and soft target updates in both environment suites.
At 5M steps, moving from gymnax to official MinAtar lowers the mean Asterix return from 14.2 to 1.4 and raises the Breakout return from 11.0 to 17.7.
On official Seaquest, the mean return is 1.1, compared with approximately 20 in Figure~3 of \citet{young2019minatar}. The published curve uses a moving average over 100 training episodes, while our returns aggregate episodes completed within each 100,000-step training epoch.

\begin{figure}[t]
  \centering
  \includegraphics[width=\linewidth]{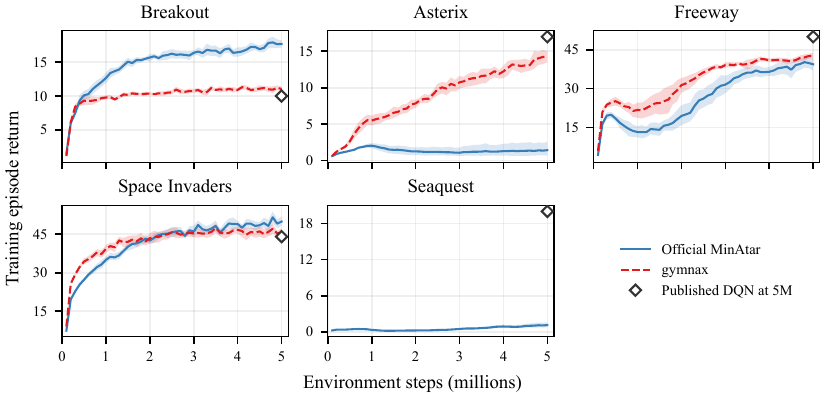}
  \caption{DQN training returns on official and gymnax MinAtar. Lines and shading show the mean and 95\% bootstrap confidence intervals over 10 seeds per game and environment suite. Diamonds mark approximate 5M-frame returns read from Figure~3 of \citet{young2019minatar}. Seaquest was run on official MinAtar.}
  \label{fig:dqn-minatar}
\end{figure}

\begin{figure}[t]
  \centering
  \includegraphics[width=\linewidth]{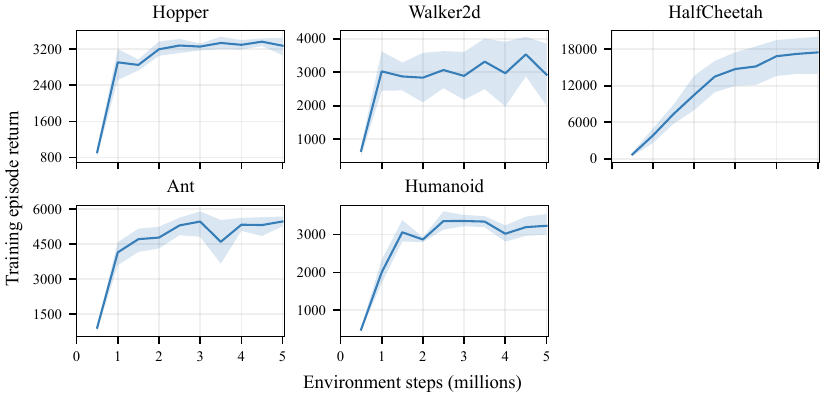}
  \caption{SAC training episode returns on five Brax MuJoCo tasks. Lines and shading show the mean and 95\% bootstrap confidence intervals over five seeds.}
  \label{fig:sac}
\end{figure}

\paragraph{Continuous control.}
Figure~\ref{fig:sac} shows SAC training returns on the Brax versions of five MuJoCo tasks.

\begin{table}[h]
  \centering
  \caption{Held-out negative log-likelihood of assistant tokens on AgentTrove for Qwen3-style models trained from scratch with recurrent BC on every token or only on the assistant's tokens, at the full estimated compute budget of each size, with 95\% bootstrap confidence intervals over 512 validation interactions. Each entry is a single run.}
  \label{tab:blm}
  \begin{tabular}{lcc}
    \toprule
    Parameters & Every token & Assistant tokens \\
    \midrule
    82M & 1.473 [1.415, 1.533] & 1.347 [1.293, 1.404] \\
    191M & 1.192 [1.141, 1.246] & 1.116 [1.068, 1.168] \\
    504M & 1.330 [1.276, 1.389] & 1.342 [1.287, 1.402] \\
    \bottomrule
  \end{tabular}
\end{table}

\clearpage
\section{Differences from related libraries}
\label{app:design}

This appendix expands on the comparison in Table~\ref{tab:comparison}.
In contrast with PureJaxRL \citep{lu2022discovered} and CleanRL \citep{huang2022cleanrl}, \boonta{} does not aim to provide one file per algorithm that can be read and forked on its own.
CleanRL includes 13 implementations of PPO, including 406 lines for multi-GPU Atari and 332 lines for single-GPU Atari. Stoix's Anakin and Sebulba PPO files contain 731 and 1046 lines, respectively, each including its own update. JaxUED implements domain randomization and prioritized level replay as training scripts of 661 and 902 lines.\footnote{Line counts use \texttt{wc -l}.}
The PPO implementation in \boonta{} is 239 lines long because it leaves the training loop, the device placement and the curriculum to other components, at the cost of having to know these interfaces to follow a complete run.
Stoix \citep{toledo2024stoix} and Mava \citep{dekock2023mava} also implement the Podracer architectures, with a broader set of single-agent algorithms and a focus on cooperative multi-agent benchmarks, respectively, while \boonta{} adds an offline podracer and the pit and lap hooks under the same interface.
PufferLib \citep{suarez2024pufferlib} focuses on fast native environments and their vectorization.
TorchRL \citep{bou2024torchrl} is, like \boonta{}, built from importable components, but targets PyTorch and relies on a shared data carrier, the TensorDict, whereas \boonta{} relies on pure functions over JAX pytrees, so that its components are compiled into a single program.
Rejax \citep{liesen2024rejax} implements each algorithm as one training function that can be compiled whole and transformed with \texttt{vmap} or \texttt{pmap}, which suits training many independent agents at once, while \boonta{} shares one policy update across devices. Jaxnasium \citep{ponse2026jaxnasium} wraps environments from many suites, including host-based Gymnasium and EnvPool environments, behind one PyTree interface, so that its algorithms act in discrete, continuous, image, composite and multi-agent spaces without changes. \boonta{} likewise runs Python environments through callbacks and native emulators through the foreign function interface (Appendix~\ref{app:environments}), and adds the podracers for distributed training.

\end{document}